\documentclass{article}
\usepackage{ijcai24}

\usepackage{times}
\usepackage{soul}
\usepackage{url}
\usepackage[hidelinks]{hyperref}
\usepackage[utf8]{inputenc}
\usepackage[small]{caption}
\usepackage{graphicx}
\usepackage{amsmath}
\usepackage{amsthm}
\usepackage{booktabs}
\usepackage{algorithm}
\usepackage{algorithmic}
\usepackage[switch]{lineno}
\usepackage{graphicx}
\usepackage{pdfpages}
\usepackage{array}

\usepackage[draft,textsize=footnotesize,textwidth=15mm]{todonotes}

\title{Parameter Efficient Fine Tuning: 
A Comprehensive Analysis Across Applications 
}

\author{
Charith Chandra Sai Balne$^1$
\and
Sreyoshi Bhaduri$^2$\footnote{Work does not relate to position at Amazon.}\and
Tamoghna Roy$^3$\footnote{Work does not relate to position at DeepSig Inc.}\And
Vinija Jain$^{4}$\And
Aman Chadha$^{4,5}$\textsuperscript{*}
\affiliations
$^1$University of Southern California\\
$^2$Amazon\\
$^3$Deepsig Inc.\\
$^4$Stanford University\\
$^5$Amazon GenAI\\
\emails
\footnotesize\texttt{{charithchandra23}@gmail.com, {sreyoshibhaduri}@gmail.com, \\{tamoghna.roy}@gmail.com, hi@vinija.ai, hi@aman.ai}
}

\begin{document}

\maketitle

\begin{abstract}
   The rise of deep learning has marked significant progress in fields such as computer vision, natural language processing, and medical imaging, primarily through the adaptation of pre-trained models for specific tasks. Traditional fine-tuning methods, involving adjustments to all parameters, face challenges due to high computational and memory demands. This has led to the development of Parameter Efficient Fine-Tuning (PEFT) techniques, which selectively update parameters to balance computational efficiency with performance. This review examines PEFT approaches, offering a detailed comparison of various strategies highlighting applications across different domains, including text generation, medical imaging, protein modeling, and speech synthesis. By assessing the effectiveness of PEFT methods in reducing computational load, speeding up training, and lowering memory usage, this paper contributes to making deep learning more accessible and adaptable, facilitating its wider application and encouraging innovation in model optimization. Ultimately, the paper aims to contribute towards insights into PEFT's evolving landscape, guiding researchers and practitioners in overcoming the limitations of conventional fine-tuning approaches.
\end{abstract}

\section{Introduction}
\label{sec:intro}

\begin{figure*}[h]
    \centering
    \includegraphics[width=0.9\textwidth]{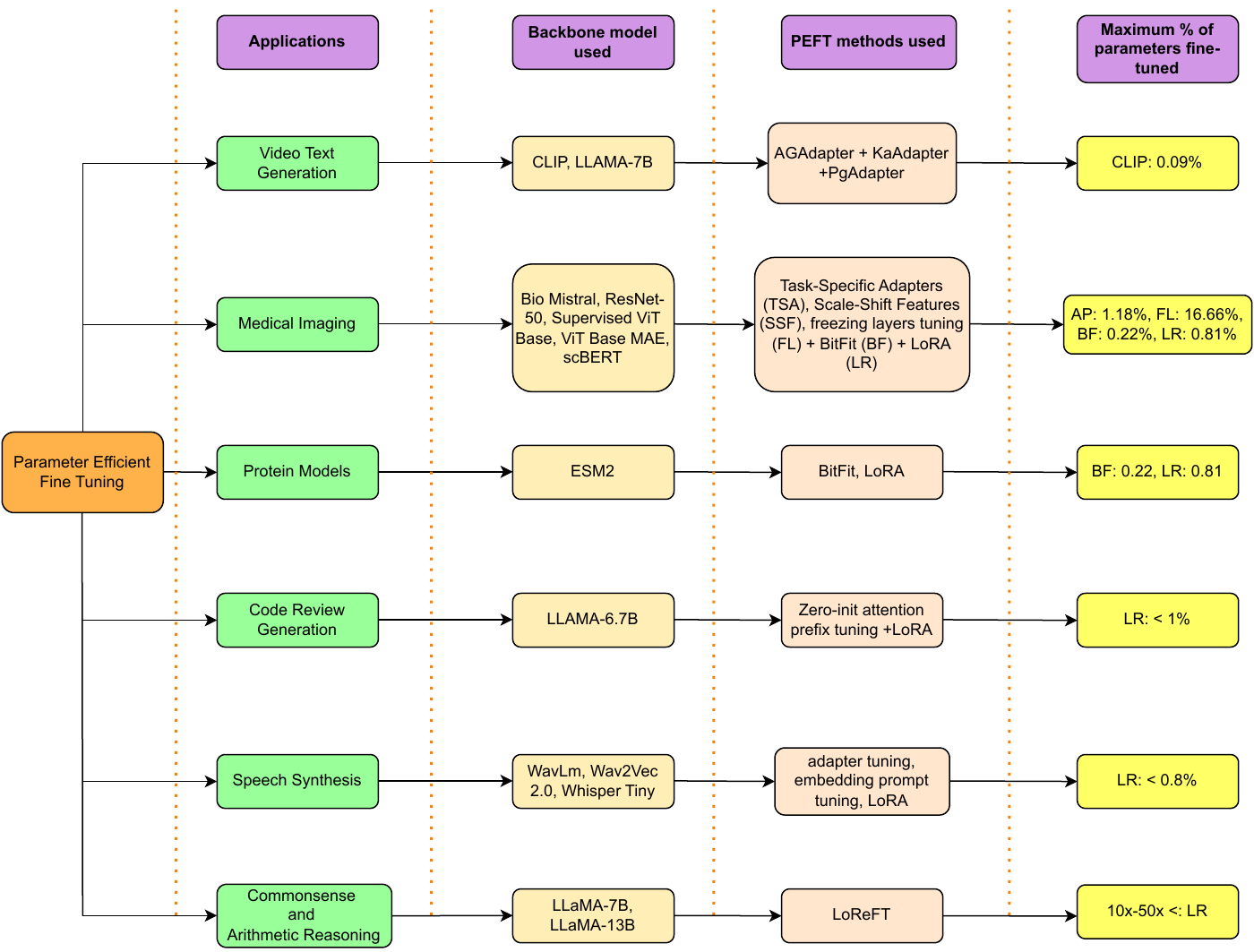}
    \caption{Comparative study of PEFT across different applications.}
    \label{fig:peft_summary}
\end{figure*}

Deep learning has revolutionized the field of artificial intelligence, enabling remarkable advancements in various applications such as Large-scale vision-language (VL) models \cite{radford2021learning}, \cite{jia2021scaling}, \cite{yao2021filip}, \cite{alayrac2022flamingo}, \cite{yuan2021florence} natural language processing \cite{lu2022imaginationaugmented}, \cite{yan2022clip}, and speech recognition \cite{nassif2019speech},\cite{prabhavalkar2023end}. However, the fine-tuning process, which involves adjusting model weights to fit new tasks or datasets, can be computationally expensive and memory-intensive. This has led to a growing interest in PEFT methods that can reduce the computational cost and memory usage while maintaining performance.

PEFT methods aim to strike a balance between accuracy and efficiency by selectively updating a subset of model parameters, leveraging knowledge distillation, or exploiting structural redundancy. These methods have the potential to significantly reduce the computational cost and memory usage, making deep learning more accessible and scalable for a wider range of applications and devices. This review paper aims to provide a comprehensive overview of the recent advances in PEFT methods, discussing their underlying principles, applications, and trade-offs. We explore state-of-the-art techniques, compare their performance, and highlight the challenges and future research directions in this emerging field. By shedding light on the efficiency aspects of fine-tuning, our paper aspires to contribute to democratizing deep learning and enabling its widespread adoption across applications.

\section{Fine-tuning Methods}

Modern pre-trained models (such as BERT \cite{devlin2018bert}, GPT {\cite{radford2019language}, T5 \cite{raffel2020exploring}, etc.) consist of billions, if not trillions (especially in case of mixture-of-experts architectures), of parameters. Traditional fine-tuning methods involve adjusting \textit{all }model parameters to fit the new task or dataset, which can be computationally expensive and memory-intensive. This approach is often referred to as "full fine-tuning" \cite{lv2023parameter}. Full fine-tuning requires a large amount of data and computational resources to converge \cite{9408054}, which can be a limitation for tasks with limited data availability or computational budgets. Additionally, fine-tuning all parameters often lead to over-fitting, especially when the new task has limited data.

Another limitation of traditional fine-tuning methods is that they do not leverage the knowledge gained during pre-training \cite{han2024parameter}. Pre-trained models are typically trained on large datasets and have learned general features that are useful across multiple tasks. Full fine-tuning discards this knowledge and starts from scratch (e.g., \cite{korbak2022controlling}), which can lead to sub-optimal performance.

Finally, traditional fine-tuning methods can result in catastrophic forgetting, where the model forgets the knowledge learned during pre-training \cite{chen2020recall}. This can lead to poor performance on both the new task and the original task, making it difficult to achieve good performance across multiple tasks. These limitations have led researchers to explore PEFT methods that can address these issues. PEFT allows to only fine-tune a small number of model parameters while freezing most of the parameters of the pre-trained LLM. PEFT has the following advantages: (i) reduced computational costs (requires fewer GPUs and GPU time); (ii) faster training times (finishes training faster); (iii) lower hardware requirements (works with cheaper GPUs with less VRAM); (iv) better modeling performance (reduces over-fitting); and (v) less storage (majority of weights can be shared across different tasks).

\section{Applications}
In this section, we explore parameter-efficient fine-tuning across various applications including  commonsense and arithmetic reasoning, generating descriptive texts for videos, enhancing medical imaging accuracy, refining protein models for better scientific insights, automating code review and generation, and advancing speech synthesis technologies. A comparative analysis of PEFT methods is given in Table 3.

\subsection{Commonsense and Arithmetic Reasoning}
Representation Fine-Tuning (ReFT) is a technique that modifies only a minimal subset of model weights to fine-tune large-scale language models through \cite{wu2024reft}. The paper presents a specific variant of ReFT, dubbed Low-rank Linear Subspace ReFT (LoReFT), which modifies the model's internal representations and exhibits far greater parameter efficiency, with improvements by factors of 10 to 50 compared to contemporary PEFT methods. The foundational mechanism of the LoReFT framework, is defined by the Distributed Interchange Intervention (DII) formula \( DII(b, s, R) = b + R^\top (Rs - Rb) \). \cite{wu2024reft} employ the projection matrix R to refine the hidden states b, steering them toward a target state s. This method is crafted to subtly yet efficiently influence the model's output, guiding it towards desired behaviors or responses. Extensive evaluations conducted by the authors on various reasoning tasks and benchmarks such as Alpaca-Eval v1.0 and GLUE indicated that LoReFT not only achieves better efficiency but also superior performance relative to leading PEFT approaches over different datasets in their respective categories.

LoReFT achieved state-of-the-art performance for commonsense reasoning, surpassing other methods such as Prefix Tuning \cite{bisk2019piqa}, Adapter-based methods, and LoRA, particularly on LLaMA-7B and LLaMA-13B models. LoReFT showed an accuracy improvement, averaging an 80.2\% and 83.3\% across different datasets BoolQ, PIQA, SIQA, HellaS., WinoG., ARC-e, ARC-c, OBQA, for the Llama 7B and 13B models respectively. See specific results from the paper in \tableautorefname{ 1}.

\begin{table}[ht]
\centering
\caption{Average performance of commonsense reasoning over BoolQ, PIQA, SIQA, HellaS., WinoG., ARC-e, ARC-c, OBQA, datasets for the LLaMA-7B and LLaMA-13B models. Comparisons from research conducted by \protect\cite{wu2024reft}}

\begin{tabular}{cccc}
\toprule
\textbf{Model} & \textbf{PEFT} & \textbf{Params (\%)} & \textbf{Avg. Accuracy} \\
\toprule
ChatGPT & — & — & 77.0\% \\
\toprule
\multicolumn{4}{l}{LLaMA-7B} \\ 
\toprule
&PrefT & 0.110\% & 64.6\% \\
&AdapterS & 0.990\% & 70.8\% \\
&AdapterP & 3.540\% & 72.3\% \\
&LoRA & 0.830\% & 74.7\% \\
&DoRA (half) & 0.430\% & 77.5\% \\
&DoRA & 0.840\% & 78.1\% \\
&LoReFT & 0.031\% & 80.2\% \\
\toprule
\multicolumn{4}{l}{LLaMA-13B} \\ 
\toprule
&PrefT & 0.030\% & 68.4\% \\
&AdapterS & 0.800\% & 79.5\% \\
&AdapterP & 2.890\% & 81.5\% \\
&LoRA & 0.670\% & 80.5\% \\
&DoRA (half) & 0.350\% & 80.8\% \\
&DoRA & 0.680\% & 81.5\% \\
&LoReFT & 0.025\% & 83.3\% \\
\toprule
\end{tabular}
\end{table}

The performance of the LoReFT in arithmetic reasoning \cite{hu-etal-2023-llm} tasks is found to be inferior to that of LoRA and adapters, though it surpasses prefix-tuning. The analysis indicates that LoReFT may encounter more challenges in chain-of-thought reasoning as opposed to single-step commonsense reasoning tasks. This difficulty is attributed to the extended length of generations, which diminishes the efficacy of the intervention, and the inherent complexity of the task. Additionally, the paper revealed that LoReFT demonstrates improved performance with the 13B model compared to the 7B model, suggesting scalability of LoReFT with increased model size.See specific results from the paper in \tableautorefname{ 2}.

\begin{table}[htbp]
\centering
\caption{Arithmetic reasoning performance of LLaMA-7B and LLaMA-13B models over AQuA, GSM8K, MAWPS, SVAMP datasets. Comparisons from research conducted by \protect\cite{wu2024reft}}
\resizebox{0.9\columnwidth}{!}{
\begin{tabular}{@{}lcccccc@{}}
\toprule
\textbf{Model} & \textbf{PEFT Params (\%)} & \textbf{AQuA} & \textbf{GSM8K} & \textbf{MAWPS} & \textbf{SVAMP} & \textbf{Avg.} \\ \midrule
\multicolumn{7}{c}{\textbf{LLaMA-7B}} \\ \midrule
PrefT          & 0.110\% & 14.2 & 24.4 & 63.4 & 38.1 & 35.0 \\
AdapterS       & 0.990\% & 15.0 & 33.3 & 77.7 & 52.3 & 44.6 \\
AdapterP       & 3.540\% & 18.1 & 35.3 & 82.4 & 49.6 & 46.4 \\
LoRA           & 0.830\% & 18.9 & 37.5 & 79.0 & 52.1 & 46.9 \\
LoReFT         & 0.031\% & 21.4 & 26.0 & 76.2 & 46.8 & 42.6 \\ \midrule
\multicolumn{7}{c}{\textbf{LLaMA-13B}} \\ \midrule
PrefT          & 0.300\% & 15.7 & 31.1 & 66.8 & 41.4 & 38.8 \\
AdapterS       & 0.800\% & 22.0 & 44.0 & 78.6 & 50.8 & 48.9 \\
AdapterP       & 2.890\% & 20.5 & 43.3 & 81.1 & 55.7 & 50.2 \\
LoRA           & 0.670\% & 18.5 & 47.5 & 83.6 & 54.6 & 51.1 \\
LoReFT          & 0.025\% & 23.6 & 38.1 & 82.4 & 54.2 & 49.6 \\ \bottomrule
\end{tabular}
}
\end{table}
\begin{table*}[ht]
\centering
\caption{Comparative analysis of prevalent PEFT methods.}
\label{tab:peft_perf}
\renewcommand{\arraystretch}{1.2}
\resizebox{0.85\textwidth}{!}{
\begin{tabular}{>{\centering\arraybackslash}m{0.3\linewidth} >{\centering\arraybackslash}m{0.3\linewidth} >{\centering\arraybackslash}m{0.3\linewidth} >{\centering\arraybackslash}m{0.3\linewidth}}
\toprule
\textbf{Method} & \textbf{Parameter reduction (\%)} & \textbf{Advantages} & \textbf{Disadvantages} \\
\midrule
Full Fine-Tuning (ViT-B/16, BARD)\par \cite{liu2022simplex} & 0 & Performant baseline & High memory footprint (33B parameters) \\ 
\midrule
Adapter Modules (Tiny)\par \cite{vandermarel2022highresolution} & 85 & Flexible, modular design & Requires hyperparameter tuning\\ 
Adapter Modules (Small) & 75 & Flexible, modular design & Requires hyperparameter tuning\\ 
\midrule
LoRA\par \cite{zhou2021topological} & 90 & Memory efficient (3.3B parameters) & Limited control over updates\\ 
\midrule

LoReFT\par \cite{wu2024reft} & 70-90 & Memory efficient, potentially interpretable & Efficiency depends on task and hyperparameters\\ 
\midrule
Prefix Tuning (Learned)\par \cite{luo2021empirical} & 65 & Simple implementation & May not capture complex video features\\ 
\midrule
Sparse Fine-Tuning (40\% pruning)\par \cite{saied2016fimodule} & 60 & Memory efficient (13.2B parameters) & Requires careful selection of parameters\\ 
Sparse Fine-Tuning (80\% pruning)\par & 80 & Extremely memory efficient (6.6B parameters) & Significant accuracy drop at high pruning ratio\\ 
\midrule
BitFit (8-bit)\par \cite{zaken2022bitfit} & 95 & Extremely memory efficient (1.65B parameters) & Limited performance gains in high-data regime\\ 
\bottomrule
\end{tabular}
}
\end{table*}

\subsection{Video Text Generation}

Video-text understanding pertains to how videos and words relate to each other. This area looks into finding videos based on text descriptions and creating captions for videos, which is key for making sense of what's happening in a video just by looking at the words linked to it. Fang et al. introduce the Alignment and Generation Adapter (AGAdapter) for enhancing video-text understanding \cite{fang2023alignment}. This integrates a knowledge-sharing alignment adapter with a large language model for video-text retrieval and video captioning tasks, achieving state-of-the-art performance on MSR-VTT and ActivityNet benchmarks. Their research introduces a novel approach to video-text understanding by integrating the pre-trained CLIP model (CLIP-bigG/14) for encoding and the LLaMA-7B model for language processing, alongside KaAdapter and Pg Adapter for efficient adaptation. These components work together within a robust tech stack that optimizes video and text alignment across various datasets, including MSR-VTT and ActivityNet, tailored with video and caption lengths set to dataset-specific requirements. Numerical results from an ablation study on the MSR-VTT dataset reveal the AGAdapter's efficacy, particularly when augmented with LIcap, showcasing remarkable enhancements in video-text retrieval and video captioning metrics compared to the CLIP-finetuned baseline. These outcomes underscore the method's success in delivering significant performance uplifts within minimal training times (0.12 to 0.5 hours), affirming its potential in advancing video-text comprehension tasks with high efficiency and effectiveness.\\

Similarly, the KAdaptation method, achieves a trade-off between accuracy and parameter efficiency in the vision transformer (ViT-B-224/32) through CLIP pretraining \cite{he2023parameterefficient}. Evaluated across 20 datasets from the ELEVATER benchmark, this approach notably excels by updating merely 0.09 percent of the model's parameters, underscoring its efficiency. This result emphasizes the method's capability to maintain high accuracy while significantly reducing the number of trainable parameters, showcasing its potential for effective and efficient model adaptation .

\subsection{Medical Imaging}
Advancements in medical imaging technologies are spearheading transformative changes across various sectors of modern medicine \cite{azizi2021big}, encompassing both clinical diagnostics and biomedical research. \cite{dutt2023parameterefficient} evaluates PEFT techniques for medical image analysis \cite{chambon2022roentgen}, \cite{kirillov2023segment}, focusing on convolutional and transformer-based networks across six datasets. It assesses 16 PEFT methods through over 600 experiments, showing performance gains of up to 22 percent in some scenarios, especially in medical text-to-image generation tasks. The study demonstrates PEFT's superiority over traditional fine-tuning in certain conditions, particularly when data is scarce or model size is large. It underscores the effectiveness of PEFT in reducing computational costs while maintaining or improving performance, making it a valuable approach for the medical domain.\cite{10385599} explore parameter-efficient fine-tuning methods for cell type annotation in scRNA-seq data using scBERT \cite{choromanski2022rethinking}. It demonstrates that such methods can achieve high performance with significantly fewer parameters. Key results show that methods like Adapter \cite{houlsby2019parameterefficient}, BitFit, and LoRA, despite reducing tunable parameters BitFit uses only 0.22 percent of the model's parameters, maintain performance close to full fine-tuning, with LoRA and a combination of BitFit and LoRA among the most effective strategies. As per the Experiment conducted FT [vanilla fine-tuning] uses 100 percent of the model's parameters, whereas parameter-efficient methods use significantly less: AP[adapter] uses 1.18 percent, FL[freezing layers tuning] uses 16.66 percent, BF[BitFit] uses 0.22 percent, and LR[LoRA] uses 0.81 percent.\\

Biomedical question answering was shown to significantly improve accuracy with only 0.152 percent of baseline parameters fine-tuned \cite{10385678}. The strategy adopted includes contrastive learning and self-consistency voting, tested on PubMedQA and BioASQ datasets. Remarkably, it achieves comparable performance to GPT-4, outperforming domain-specific models without external knowledge. The T5 models highlights efficient tuning in resource-constrained environments, balancing performance and computational costs.

\subsection{Protein Models}
Large-scale protein models have significantly transformed the field of proteomics through their capacity to learn from extensive volumes of sequence data autonomously. Later, these models get a bit of training on specific tasks to make them even better at what they do \cite{Sledzieski2023.11.09.566187} introduced parameter-efficient fine-tuning methods for protein language models, focusing on tasks like protein-protein interaction (PPI) prediction and homooligomer symmetry prediction. It shows that PEFT can achieve comparable or superior performance to traditional fine-tuning with significantly fewer parameters. For PPI prediction, PEFT models even outperform traditional methods. Despite the dramatic reduction in tunable parameters (BitFit at 0.22 percent , Adapter at 1.18 percent, Low-Rank Adaptation at 0.81 percent, and Freezing Layers at 16.66 percent compared to the full model's 100 percent), these methods maintain or nearly match the performance of traditional fine-tuning across various datasets. For instance, on the Zheng68k dataset, accuracy and F1 scores were closely aligned across methods, with Adapter and Low-Rank Adaptation showing particularly strong performance. Similar trends were observed in the Baron-human and Baron-mus datasets, where these parameter-efficient methods achieved high accuracy and F1 scores, showcasing their capability to deliver efficient and scalable solutions for cell type annotation while significantly reducing computational resources.
\begin{figure*}[h]
    \centering
    \includegraphics[width=0.9\textwidth]{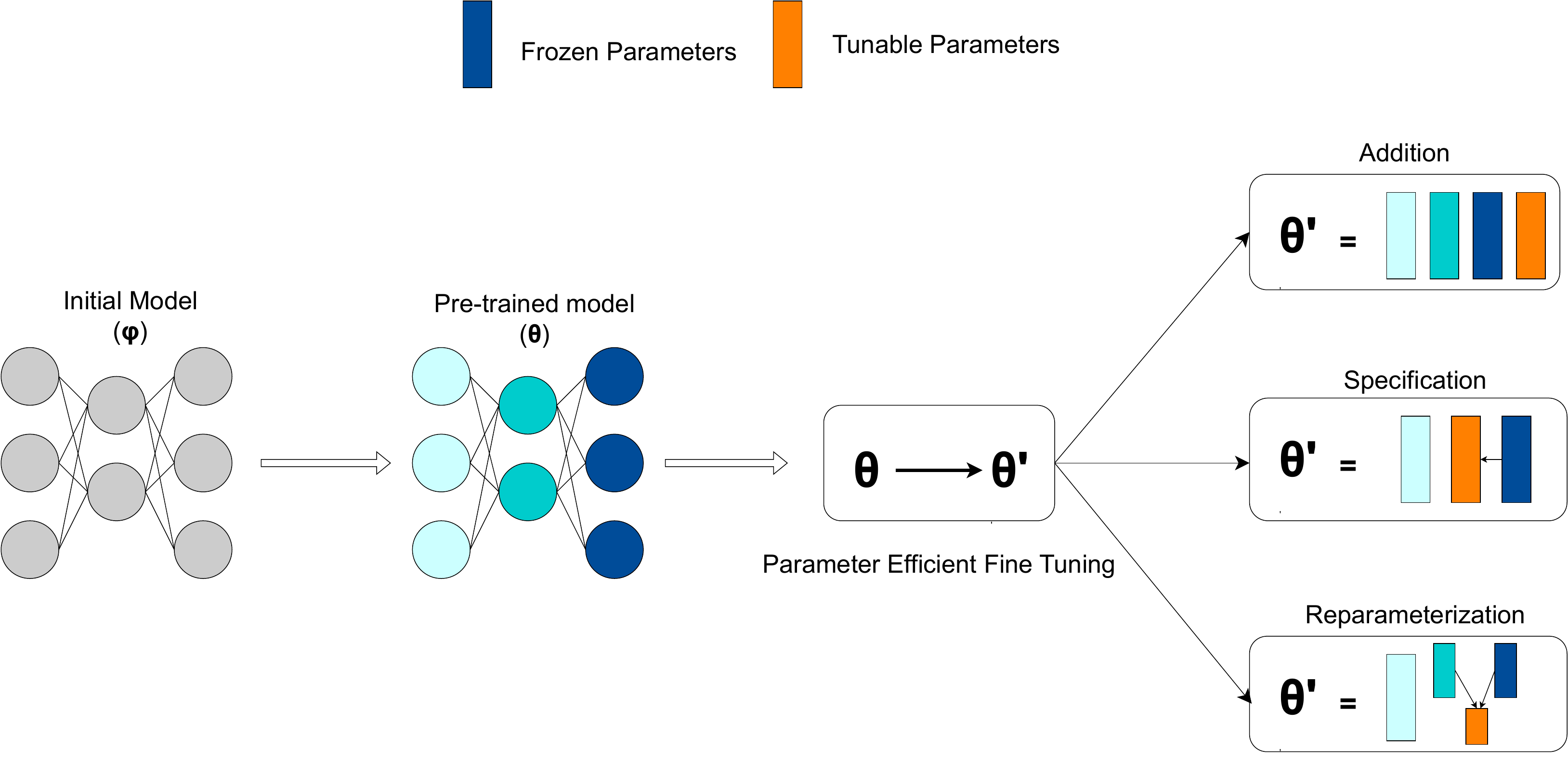}
    \caption{Illustration of workflow for the PEFT paradigm starting with a pre-trained model ($\theta$), to which modifications such as additions, specifications, and reparameterizations are applied, effectively differentiating between frozen and tunable parameters to enhance model performance.}
    \label{fig:peft_workflow}
\end{figure*}

\subsection{Code Review / Generation}
Since Fagan \cite{fang2023alignment} introduced it in 1976, code review has been key in finding bugs, improving quality, and sharing knowledge in software development. But, this mostly manual task can really pile on the work for developers. Even with today's modern code review methods, which are a bit smoother than the old ways, it still asks a lot from them. \cite{lu2023llamareviewer} The study introduces LLaMA-Reviewer, a framework that automates code review tasks by leveraging PEFT techniques on the LLaMA model. It achieved notable numerical insights across various metrics: For Review Necessity Prediction on the CRer dataset, it reached a precision of 60.99 percent, a recall of 83.50 percent, and an F1 score of 70.49 percent using Low-Rank Adaptation (LoRA). In Code Review Comment Generation, LLaMA-Reviewer scored BLEU-4 scores of 5.70 on the CRer dataset and 5.04 on the Tufano dataset, showcasing its superior performance over existing models like CodeReviewer and AUGER. Additionally, for Code Refinement tasks, it attained BLEU-4 scores of 82.27 on the CRer dataset and 78.23 on the Tufano dataset, demonstrating its competitive or superior capability compared to traditional models. These results highlight LLaMA-Reviewer's efficiency in code review automation, offering promising directions for future software engineering research with a focus on minimizing the need for extensive parameter tuning while maintaining high performance.
\subsection{3D Pretrained Models}
In exploring efficient approaches for fine-tuning pre-trained 3D models, a novel framework named Point-PEFT \cite{tang2023pointpeft} has been proposed, demonstrating enhanced performance over traditional full fine-tuning methods with a significantly reduced computational footprint. Notably, Point-PEFT managed to outperform the full fine-tuning benchmarks on ModelNet40 and ScanObjectNN \cite{uy2019revisiting}, achieving accuracy levels of 94.2\% and 89.1\% respectively, while requiring merely 5\% of the trainable parameters compared to 22.1M parameters in the full fine-tuning setup. Such results underscore the efficiency and general applicability of Point-PEFT across various pre-trained 3D models, including Point-BERT \cite{yu2022pointbert} and Point-M2AE \cite{zhang2022pointm2ae}, highlighting its potential for broader adoption in the field of 3D point cloud processing \cite{tang2024pointpeft}

\subsection{Speech Synthesis}

In \cite{Feng_2023}, the authors meticulously evaluated the effectiveness of PEFT methods, namely adapter tuning, embedding prompt tuning, and Low-rank approximation (LoRA), across four prominent SER \cite{chen2023exploring},\cite{feng2023trustser}datasets \cite{houlsby2019parameterefficient}. Fine-tuning methods comparatively provided better results than previous methods, which were solely dependent on MLP (Multilayer Perceptron), CNN (Convolutional Neural Networks), RNN (Recurrent Neural Networks), Mixed data Neural Networks \cite{9691563} by extracting higher-order melfrequency cepstral coefficients \cite{6967289}. The results reveal a notable superiority of LoRA in enhancing the fine-tuning performance of pre-trained speech models for emotion recognition tasks by using generative \cite{Chen_2022}, dis-
criminative \cite{baevski2020wav2vec}, \cite{schneider2019wav2vec} and multi-task learning objectives. Specifically, LoRA outperformed other PEFT methods, achieving the highest average Unweighted Average Recall (UAR) of 67.3\% on the WavLM Base+ model, demonstrating its effectiveness in adapting pre-trained models to SER tasks efficiently. In contrast, traditional adapter tuning and embedding prompt methods yielded lower performance, with adapter tuning achieving an average UAR of 63.07`\%` on the Wav2Vec 2.0 Base model \cite{radford2022robust} and embedding prompt tuning showing less impact on performance across various models. Furthermore, the study highlighted the minimal additional parameter requirement introduced by LoRA, underlining its practicality for real-world applications. Additionally, the research underscored the importance of fairness in SER systems, with LoRA showing promising results in improving fairness scores across multiple datasets. These findings not only demonstrate the potential of LoRA in achieving high performance and fairness in SER tasks but also pave the way for future research directions focusing on the optimization of PEFT methods for speech emotion recognition. A similar and innovative study in \cite{liu2024sparsely} states child whisper recognition, whereas \cite{9977793} uses some similar techniques of transfer learning to understand child behaviours using their speech and cry sounds.

\section{Considerations for Evaluation Across PEFT Methods}

PEFT has emerged as a compelling approach for tailoring large pre-trained models to specific tasks while minimizing computational demands. Our review found that leveraging PEFT across diverse applications presents several key challenges that require careful consideration, as practitioners consider applying PEFT for their applications:

\textbf{A) Balancing Efficiency and Performance}: A core challenge lies in striking a delicate balance between reducing trainable parameters and maintaining robust performance \cite{naveed2024comprehensive}. Fine-tuning too few parameters might hinder the model's ability to adapt effectively to the target task, while excessively fine-tuning can negate the computational benefits of PEFT\cite{dutt2023parameterefficient}.

\textbf{B) Data Scarcity and Generalizability}: The success of PEFT can be contingent on the quality and quantity of data available for fine-tuning. In domains with limited or noisy data, PEFT may struggle to achieve the same level of accuracy attainable with full fine-tuning on a larger dataset \cite{dutt2024fairtune}. Careful selection of data augmentation techniques and transfer learning strategies\cite{9977793} can be crucial to mitigate this challenge.

\textbf{C) Over-fitting and Generalization Trade-off}: There is an inherent risk of over-fitting the model to the training data \cite{chavan2024faster}, particularly when using a restricted set of parameters for fine-tuning. This can lead to a scenario where the model performs well on the training data but exhibits poor performance on unseen examples. To address this, employing appropriate regularization techniques and meticulous hyperparameter tuning becomes essential to promote better generalization to new data \cite{kirk2024understanding}.

\textbf{D) Capacity Constraints of Incremental Modules}: Certain PEFT methods introduce additional modules with a reduced number of parameters on top of the pre-trained model. The challenge here lies in ensuring that these smaller modules possess sufficient capacity to learn the intricacies of the specific task effectively, especially when there are strict constraints on the allowable number of parameters. Ongoing research is focused on developing methods to enhance the capacity of these modules without compromising parameter efficiency.

\section{Discussions}
This study provides an exhaustive review of the literature concerning the effectiveness of various PEFT techniques across multiple applications. 

These include Video Text Generation utilizing distinct adaptors for downstream tasks, Biomedical Imaging characterized by stringent data confidentiality and significant annotation costs, Protein models necessitating extensive parameters for comprehensive fine-tuning, and Code Review Generation. Our analysis reveals that Low-Rank Adaptation (LoRA) fine-tunes a minimal number of parameters, thus enabling the recalibration of training weights on a single GPU. Conversely, Differentiable Rank Adaptation (DoRA) demonstrates superior performance, outperforming LoRA.

We also propose several potential directions for future research to further advance the PEFT field, particularly focusing on the evaluation of specific applications:

\textbf{A) Task-Agnostic PEFT Techniques:}

Future research should focus on developing PEFT methods that are universally applicable across different downstream tasks. This would reduce the necessity for specialized adaptors in each application domain, enhancing the flexibility and ease of PEFT deployment. Exploring meta-learning or transferable parameter approaches may achieve task-agnostic efficacy.

\textbf{B) Privacy-Preserving PEFT for Sensitive Data:}

In fields such as biomedical imaging where data privacy is crucial, it is essential to adapt PEFT to operate on sensitive datasets without breaching patient confidentiality. Exploring federated learning or homomorphic encryption techniques could allow for privacy-preserving PEFT.

\textbf{C) Limited Labeled Data and PEFT:}

Given the frequent scarcity of labeled data in domains like biomedical imaging, enhancing the robustness of PEFT in these contexts is critical. Future investigations could consider active learning or curriculum learning techniques to improve fine-tuning under limited data conditions.

\textbf{D) Interpretability of Fine-Tuned Protein Models:}

While PEFT reduces the parameter count in protein models, its impact on model interpretability remains uncertain. Future research should examine methods to elucidate the decision-making processes and mechanisms within these fine-tuned models.







By addressing these future research directions, we can fully harness the capabilities of PEFT, ensuring its progressive development for efficient and effective fine-tuning of large models across diverse applications.

\appendix

\bibliographystyle{named}
\bibliography{ijcai24}

\end{document}